%% file: acl2020.tex
\documentclass[11pt,a4paper]{article}
\usepackage[hyperref]{acl2020}
\usepackage{times}
\usepackage{latexsym}

\usepackage{hyperref}
\usepackage{microtype}

\usepackage{multirow}
\usepackage{amsmath,amssymb, stmaryrd}
\usepackage{pgf}

\usepackage{mathtools}
\DeclarePairedDelimiter{\opair}{\langle}{\rangle}

\usepackage{array}
\newcolumntype{P}[1]{>{\centering\arraybackslash}p{#1}}

\usepackage{xcolor, color, colortbl, pifont}
\usepackage{graphicx}
\usepackage{subcaption}

\usepackage{todonotes}
\usepackage{booktabs}
\usepackage{url}
\usepackage{fixltx2e}
\usepackage{verbatim}
\usepackage{pgfplots}
\usepgfplotslibrary{groupplots}
\pgfplotsset{compat = 1.11}
\pgfplotsset{width=7cm,compat=1.8}

\definecolor{grannysmithapple}{rgb}{0.66, 0.89, 0.63}
\definecolor{green(munsell)}{rgb}{0.0, 0.66, 0.47}
\definecolor{green(ncs)}{rgb}{0.0, 0.62, 0.42}
\definecolor{huntergreen}{rgb}{0.21, 0.37, 0.23}

\definecolor{pastelred}{rgb}{1.0, 0.41, 0.38}
\definecolor{red(munsell)}{rgb}{0.95, 0.0, 0.24}
\definecolor{red(ncs)}{rgb}{0.77, 0.01, 0.2}
\definecolor{rosewood}{rgb}{0.4, 0.0, 0.04}

\definecolor{red-brown}{rgb}{0.65, 0.16, 0.16} % unused
\definecolor{ghostwhite}{rgb}{0.97, 0.97, 1.0}

\usepackage{tikz}
\usetikzlibrary{shapes.arrows}
\newcommand{\FancyUpArrowLight}{\begin{tikzpicture}[baseline=-0.3em]
\node[single arrow,draw,rotate=90,single arrow head extend=0.4em,inner
ysep=0.4em,transform shape,line width=0.0em, fill=grannysmithapple] (X){};
\end{tikzpicture}}
\newcommand{\FancyUpArrowMedium}{\begin{tikzpicture}[baseline=-0.3em]
\node[single arrow,draw,rotate=90,single arrow head extend=0.4em,inner
ysep=0.4em,transform shape,line width=0.0em, fill=green(munsell)] (X){};
\end{tikzpicture}}

\newcommand{\FancyUpArrowNeutral}{\begin{tikzpicture}[baseline=-0.3em]
\node[single arrow,draw,rotate=90,single arrow head extend=0.4em,inner
ysep=0.4em,transform shape,line width=0.0em, fill=ghostwhite] (X){};
\end{tikzpicture}}

\newcommand{\FancyDownArrowMedium}{\begin{tikzpicture}[baseline=-0.6em]
\node[single arrow,draw,rotate=-90,single arrow head extend=0.4em,inner
ysep=0.4em,transform shape,line width=0.0em, fill=red(ncs)] (X){};
\end{tikzpicture}}

\aclfinalcopy % Uncomment this line for the final submission
\title{Mind the Trade-off: Debiasing NLU Models without \\Degrading the In-distribution Performance}

\author{Prasetya Ajie Utama$^{\dag\ddag}$ , Nafise Sadat Moosavi$^{\ddag}$, Iryna Gurevych$^{\ddag}$\\
  \\
  $^{\dag}$Research Training Group AIPHES\\
  $^{\ddag}$Ubiquitous Knowledge Processing Lab (UKP-TUDA)\\
  Department of Computer Science, Technische Universit{\"a}t Darmstadt\\
  \url{https://www.ukp.tu-darmstadt.de}
  }

\date{}

\begin{document}
\maketitle
\begin{abstract}
Models for natural language understanding (NLU) tasks often rely on the idiosyncratic biases of the dataset, which make them brittle against test cases outside the training distribution. Recently, several proposed debiasing methods are shown to be very effective in improving out-of-distribution performance. However, their improvements come at the expense of performance drop when models are evaluated on the in-distribution data, which contain examples with higher diversity. 
This seemingly inevitable trade-off may not tell us much about the changes in the reasoning and understanding capabilities of the resulting models on broader types of examples beyond the small subset represented in the out-of-distribution data. In this paper, we address this trade-off by introducing a novel debiasing method, called \emph{confidence regularization}, which discourage models from exploiting biases while enabling them to receive enough incentive to learn from all the training examples. We evaluate our method on three NLU tasks and show that, in contrast to its predecessors, it improves the performance on out-of-distribution datasets (e.g., 7pp gain on HANS dataset) while maintaining the original in-distribution accuracy.\footnote{The code is available at \url{https://github.com/UKPLab/acl2020-confidence-regularization}}

\end{abstract}

\input{intro.tex}
\input{related.tex}
\input{method.tex}
\input{exp.tex}
\input{discussion.tex}

\section{Conclusion}

Existing debiasing methods improve the performance of NLU models on out-of-distribution datasets. However, this improvement comes at the cost of strongly diminishing the training signal from a subset of the original dataset, which in turn reduces the in-distribution accuracy. In this paper, we address this issue by introducing a novel method that regularizes models' confidence on biased examples. This method allows models to still learn from all training examples without exploiting the biases. Our experiments on four out-of-distribution datasets across three NLU tasks show that our method provides a competitive out-of-distribution performance while preserves the original accuracy. 

Our debiasing framework is general and can be extended to other task setups where the biases leveraged by models are correctly identified. Several challenges in this direction of research may include extending the debiasing methods to overcome multiple biases at once or to automatically identify the format of those biases which simulate a setting where the prior knowledge is unavailable.

\section*{Acknowledgments}
We thank Leonardo Ribeiro and Max Glockner for the thoughtful discussion on the earlier version of this work and the anonymous reviewers for their constructive comments. We also thank Tal Schuster for the support in using the Fever-Symmetric dataset. This work is supported by the German Research Foundation through the research training group “Adaptive Preparation of Information from
Heterogeneous Sources” (AIPHES, GRK 1994/1) and by the German Federal Ministry of Education and Research and the Hessian State Ministry for Higher Education, Research and the Arts within their joint support of the National Research Center for Applied Cybersecurity ATHENE.

% The acknowledgments should go immediately before the references. Do not number the acknowledgments section.
% Do not include this section when submitting your paper for review.

\bibliography{acl2020}
\bibliographystyle{acl_natbib}

\clearpage

\appendix

\input{appendix.tex}

\end{document}

%% file: intro.tex
\section{Introduction}
Despite the impressive performance on many natural language understanding (NLU) benchmarks \cite{wang2018glue}, recent pre-trained language models (LM) such as BERT \cite{devlin2018bert} are shown to rely heavily on idiosyncratic biases of datasets \cite{McCoy2019RightFT, schuster2019towards, Zhang2019PAWSPA}. These biases are commonly characterized as \textit{surface features} of input examples that are strongly associated with the target labels, e.g., occurrences of negation words in natural language inference (NLI) datasets which are biased towards the \textit{contradiction} label \cite{Gururangan2018AnnotationAI, Poliak2018HypothesisOB}. As a ramification of relying on biases, models break on the \emph{out-of-distribution} data, in which such associative patterns between the surface features and the target labels are not present. This brittleness has, in turn, limited their practical applicability in some extrinsic use cases \cite{falke-etal-2019-ranking}.

This problem has sparked interest among researchers in building models that are robust against \emph{dataset biases}. Proposed methods in this direction build on previous works, which have largely explored the format of several prominent label-revealing biases on certain datasets \cite{belinkovAdv2019}. Two current prevailing methods, \emph{product-of-expert} \cite{He2019UnlearnDB, Mahabadi2019SimpleBE} and \emph{learned-mixin} \cite{Clark2019DontTT} introduce several strategies to overcome the \emph{known} biases by correcting the conditional distribution of the target labels given the presence of biased features. They achieve this by reducing the importance of examples that can be predicted correctly by using only biased features. As a result, models are forced to learn from harder examples
in which utilizing solely superficial features is not sufficient to make correct predictions.

While these two state-of-the-art debiasing methods provide a remarkable improvement on the targeted out-of-distribution test sets, they do so at the cost of degrading the model's performance on the \emph{in-distribution} setting, i.e., evaluation on the original test data which contains more diverse inference phenomena. It raises a question on whether these debiasing methods truly help in capturing a better notion of language understanding or simply biasing models to other directions. Ideally, if such an improvement is achieved for the right reasons (i.e., better reasoning capabilities by learning a more general feature representation), a debiased model should still be able to maintain its accuracy on previously unambiguous instances (i.e., instances that are predicted correctly by the baseline model), even when they contain biases.

In this work, we address this shortcoming by introducing a novel debiasing method that improves models' performance on the out-of-distribution examples while preserves the in-distribution accuracy. The method, called \emph{confidence regularization}, draws a connection between the robustness against dataset biases and the overconfidence prediction problem in neural network models \cite{Feng2018PathologiesON, Papernot2015DistillationAA}. We show that by preventing models from being overconfident on biased examples, they are less likely to exploit the simple cues from these examples. The motivation of our proposed training objective is to \emph{explicitly} encourage models to make predictions with lower \emph{confidence} (i.e., assigning a lower probability to the predicted label) on examples that contain biased features.

\begin{table}
    \centering
    \setlength{\tabcolsep}{3pt}
    \resizebox{\columnwidth}{!}{%
    \begin{tabular}{r P{2cm} | P{2cm} | P{2cm}}
        \toprule
           & product-of-expert & learned-mixin & \textbf{conf-reg (our)}\\
        \midrule
        %  & \textit{in} & \textit{out} & \textit{in} & \textit{out} & \textit{in} & \textit{out} \\
        % \midrule
        % MNLI \textsubscript{lexical-overlap} & \FancyDownArrowMedium & \FancyUpArrowHeavy & \FancyDownArrowNeutral & \FancyUpArrowLight & \FancyUpArrowLight & \FancyUpArrowHeavy \\
        % MNLI \textsubscript{hypothesis-only} & \FancyDownArrowMedium & \FancyUpArrowMedium & \FancyDownArrowHeavy & \FancyUpArrowHeavy & \FancyUpArrowLight & \FancyUpArrowMedium \\
        % Fever \textsubscript{claim-only} & \FancyDownArrowMedium & \FancyUpArrowMedium & \FancyDownArrowHeavy & \FancyUpArrowLight & \FancyUpArrowHeavy & \FancyUpArrowMedium \\
        % QQP \textsubscript{lexical-overlap} & \FancyDownArrowNeutral & \FancyUpArrowHeavy & \FancyDownArrowNeutral & \FancyUpArrowHeavy & \FancyDownArrowNeutral & \FancyUpArrowLight \\
        in-distribution & \FancyDownArrowMedium & \FancyDownArrowMedium & \FancyUpArrowLight \\
        out-of-distribution & \FancyUpArrowMedium & \FancyUpArrowMedium & \FancyUpArrowMedium \\
        calibration & \FancyUpArrowNeutral & \FancyDownArrowMedium & \FancyUpArrowLight \\
        \midrule
        requires biased model & \textcolor{black}{\ding{52}} & \textcolor{black}{\ding{52}} & \textcolor{black}{\ding{52}} \\
        requires hyperparameter & \textcolor{green(munsell)}{\ding{54}} & \textcolor{red(ncs)}{\ding{52}} & \textcolor{green(munsell)}{\ding{54}} \\
        \bottomrule
    \end{tabular}}
    \caption{Comparison of our method against the state-of-the-art debiasing methods. Learned-mixin \cite{Clark2019DontTT} is a parameterized variant of Product-of-expert \cite{He2019UnlearnDB, Mahabadi2019SimpleBE}. Our novel confidence regularization method improves the out-of-distribution performance while optimally maintain the in-distribution accuracy.} 
    \label{tab:comparison}
    % \vspace{-4mm}
\end{table}

Table~\ref{tab:comparison} shows the comparison of our method with the existing state-of-the-art debiasing methods: \emph{product-of-expert} and \emph{learned-mixin}. We show that our method is highly effective in improving out-of-distribution performance while preserving the in-distribution accuracy. For example, our method achieves 7 points gain on an out-of-distribution NLI evaluation set, while slightly improves the in-distribution accuracy. Besides, we show that our method is able to improve models' calibration \cite{Guo2017OnCO} so that the confidences of their predictions are more aligned with their accuracies.
Overall, our contributions are the following:
\begin{itemize}
    \item We present a novel \textit{confidence regularization} method to prevent models from utilizing biased features in the dataset. We evaluate the advantage of our method over the state-of-the-art debiasing methods on three tasks, including natural language inference, fact verification, and paraphrase identification. Experimental results show that our method provides competitive out-of-distribution improvement while retaining the original in-distribution performance.
    \item We provide insights on how the debiasing methods behave across different datasets with varying degrees of biases and show that our method is more optimal when enough bias-free examples are available in the dataset.
\end{itemize}

%% file: related.tex
\section{Related Work}
\paragraph{Biases in Datasets} 
Researchers have recently studied more closely the success of large fine-tuned LMs in many NLU tasks and found that models are simply better in leveraging biased patterns instead of capturing a better notion of language understanding for the intended task \cite{bender2020climbing}. Models' performance often drops to a random baseline when evaluated on out-of-distribution datasets which are carefully designed to be void of the biases found in the training data. Using such targeted evaluation, \citet{McCoy2019RightFT} observe that models trained on MNLI dataset \cite{Williams2018ABC} leverage syntactic patterns involving word overlap to blindly predict entailment. Similarly, \citet{schuster2019towards} show that the predictions of fact verification models trained for the FEVER task \cite{Thorne2018TheFE} are largely driven by the presence of indicative words in the input claim sentences.

Following similar observations across other tasks and domains, e.g., visual question-answering \cite{agrawal-etal-2016-analyzing}, paraphrase identification \cite{Zhang2019PAWSPA}, and argument reasoning comprehension \cite{niven2019probing}, researchers proposed improved data collection techniques to reduce the artifacts that result in dataset biases. While these approaches are promising, only applying them without additional efforts in the modeling part may still deliver an unsatisfactory outcome. For instance, collecting new examples by asking human annotators to conform to specific rules may be costly and thus limit the scale and diversity of the resulting data \cite{Kaushik2020Learning}. Recently proposed adversarial filtering methods \cite{Zellers2019HellaSwagCA, winogrande2019} are more cost effective but are not guaranteed to be artifacts-free.
 It is, therefore, crucial to develop learning methods that can overcome biases as a complement to the data collection efforts. 

\paragraph{Debiasing Models}
There exist several methods that aim to improve models' robustness and generalization by leveraging the insights from previous work about the datasets' artifacts.
In the NLI task, \citet{belinkovAdv2019} make use of the finding that partial input information from the hypothesis sentence is sufficient to achieve reasonable accuracy. They then remove this hypothesis-only bias from the input representation using an adversarial training technique.
% \citet{belinkov-etal-2019-dont} propose an adversarial training method to prevent NLI models from relying only on the hypothesis-only features. 
More recently, three concurrent works \cite{Clark2019DontTT, He2019UnlearnDB, Mahabadi2019SimpleBE} introduce a model-agnostic debiasing method for NLU tasks called \texttt{product-of-expert}. \citet{Clark2019DontTT} also propose an adaptive variant of this method called \texttt{learned-mixin}. These two methods first identify examples that can be predicted correctly based only on biased features. This step is done by using a \emph{biased model}\footnote{We follow the terminology used by \citet{He2019UnlearnDB}.}, which is a weak classifier that is trained using only features that are known to be insufficient to perform the task but  work well due to biases. 
The output of this pre-trained biased model is then used to adjust the loss function such that it down-weights the importance of examples that the biased model can solve. While this approach prevents models from learning the task mainly using biased features, it also reduces model's ability to learn from examples that can be solved using these features. As a result, models are unable to optimize accuracy on the original training distribution, and they possibly become biased in some other ways.

Similar to these methods, our method also uses a biased model to identify examples that exhibit biased features. However, instead of using it to diminish the training signal from these examples, we use it to scale the confidence of models' predictions. This enables the model to receive enough incentive to learn from all of the training examples.

\paragraph{Confidence Regularization} Methods for regularizing the output distribution of neural network models have been used to improve generalization. \citet{Pereyra2017RegularizingNN} propose to penalize the entropy of the output distribution for encouraging models to be less confident in their predictions. Previously, \citet{Szegedy2015RethinkingTI} introduce a label smoothing mechanism to reduce overfitting by preventing the model from assigning a full probability to each training example. 
Our method regularizes models' confidence differently: we first perform an adaptive label smoothing for the training using knowledge distillation \cite{Hinton2015DistillingTK}, which, by itself, is known to improve the overall performance. However, our method involves an additional bias-weighted scaling mechanism within the distillation pipelines. As we will show, our proposed scaling mechanism is crucial in leveraging the knowledge distillation technique for the purpose of overcoming the targeted bias while maintaining high accuracy in the training distribution.

Similar to our work, \citet{Feng2018PathologiesON} propose a regularization method that encourages the model to be uncertain on specific examples. However, the objective and the methodology are different: they apply an entropy penalty term on examples that appear nonsensical to humans with the goal of improving models' interpretability. On the contrary, we apply our confidence regularization on every training example with a varying strength (i.e., higher uncertainty on more biased examples) to improve models' performance on the out-of-distribution data.

%% file: method.tex
\begin{figure*}[htb]
    \centering
    \includegraphics[width=15cm]{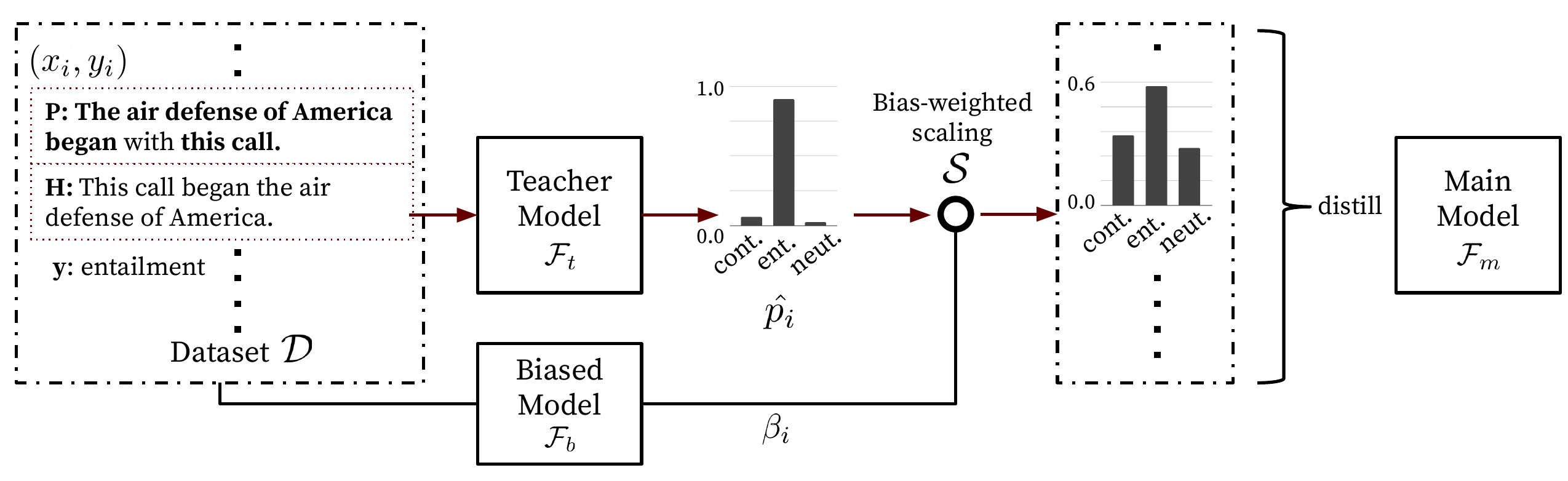}
    \caption{An overview of our debiasing strategy when applied to the MNLI dataset. An input example that contains lexical-overlap bias is predicted as entailment by the teacher model with a high confidence. When biased model predicts this example well, the output distribution of the teacher will be re-scaled to indicate higher uncertainty (lower confidence). The re-scaled output distributions are then used to distill the main model.}
    \label{fig:method_illustration}
    % \vspace{-4mm}
\end{figure*}
\section{Method}
\label{sec:method}
\paragraph{Overview} We consider the common formulation of NLU tasks as a multi-class classification problem. Given a dataset $\mathcal{D}$ that consists of $n$ examples $(x_i, y_i)_{i\in[1,n]}$, with $x_i \in \mathcal{X}$ as a pair of sentences, and $y_i \in \{1,2,...,K\}$ where $K$ is the number of classes. 
The goal is to learn a robust classifier $\mathcal{F}_m$, which computes the probability distribution over target labels, i.e., $\mathcal{F}_m(x_i) = p_i$.

The key idea of our method is to \textit{explicitly} train $\mathcal{F}_m$ to compute \emph{lower probability}, i.e., less confidence, on the predicted label when the input example exhibits a bias. This form of confidence regularization can be done by computing the loss function with the ``soft'' target labels that are obtained through our proposed smoothing mechanism. The use of soft targets as the training objective is motivated by the observation that the probability distribution of labels for each sample provides valuable information about the underlying task \cite{Hinton2015DistillingTK, Pereyra2017RegularizingNN}. When the soft targets of certain examples have higher entropy, models can be explicitly taught that some labels are more likely to be correct than the others. Based on this intuition, we argue that adjusting the confidence on soft labels can better inform the model about the true conditional distribution of the labels given the presence of the biased features.

We first produce a meaningful softened target distribution for each training example by performing \emph{knowledge distillation} \cite{Hinton2015DistillingTK}. 
In this learning framework, a ``teacher'' model $\mathcal{F}_t$, which we parameterize identically to the main model $\mathcal{F}_m$, is trained on the dataset $\mathcal{D}$ using a standard classification loss. We then use $\mathcal{F}_t$ to compute output probability distribution $\hat{p_i}$, where $\mathcal{F}_t(x_i)=\hat{p_i}$. 
In the original knowledge distillation approach, the output of the teacher model $\hat{p_i}$ is then used to train $\mathcal{F}_m$. We extend this approach by adding a novel scaling procedure before we distill the teacher model into $\mathcal{F}_m$. We define a scaling function $\mathcal{S}$ that takes the probability distribution $\hat{p_i}$ and scale it such that the probability assigned to its predicted label is lowered when the example can be predicted well by only relying on the biased features.

\paragraph{Training the biased model} For several NLU tasks, biased features are known a-priori, e.g., the word overlapping features in NLI datasets are highly correlated with the \textit{entailment} label \cite{McCoy2019RightFT}. We leverage this a-priori knowledge to design a measure of how well an example can be predicted given only the biased features. We refer to this measure as \emph{bias weight}, denoted as $\beta_i$ for every example $x_i$.

Similar to previous debiasing methods \cite{Clark2019DontTT}, we compute bias weights using a \emph{biased model}. This biased model, denoted as $\mathcal{F}_b$, predicts the probability distribution $b_i$, where $\mathcal{F}_b(x_i) = b_i = \opair{b_{i,1}, b_{i,2},...,b_{i,K}}$. We define the bias weight $\beta_i$ as the scalar value of the assigned probability by $\mathcal{F}_b$ to the ground truth label: $\beta_i=b_{i,c}$ ($c$-th label is the ground truth). 

\paragraph{Bias-weighted scaling} As illustrated in Figure~\ref{fig:method_illustration}, our method involves scaling the teacher output $\hat{p_i}$ using $\beta_i$. We do this by defining a scaling function $\mathcal{S}:\mathbb{R}^K \rightarrow \mathbb{R}^K$:
$$
   \mathcal{S}(\hat{p_i}, \beta_i)_j = \frac{\hat{p_{i,j}}^{(1-\beta_i)}}{\sum_{k=1}^K \hat{p_{i,k}}^{(1-\beta_i)}}
$$
for $j = 1, ..., K$. The value of $\beta_i$ controls the strength of the scaling: as $\beta_i \rightarrow 1$, the scaled probability assigned to each label approaches $\frac{1}{K}$, which presents a minimum confidence. Conversely, when $\beta_i \rightarrow 0$, the teacher's probability distribution remains unchanged, i.e., $\mathcal{S}(\hat{p_i}, 0) = \hat{p_i}$. 

\paragraph{Training the main model} The final step is to train $\mathcal{F}_m$ by distilling from the scaled teacher model's outputs. Since the main model is parameterized identically to the teacher model, we refer to this step as self-distillation \cite{Furlanello2018BornAgainNN}.  Self-distillation is performed by training $\mathcal{F}_m$ on pairs of input and the obtained soft target labels $(x_i, \mathcal{S}(\hat{p_i}, \beta_i))$. Specifically, $\mathcal{F}_m$ is learned by minimizing a standard cross-entropy loss between the scaled teacher's output $\mathcal{S}(\hat{p_i}, \beta_i)$ and the current prediction of the main model:
$$
    \mathcal{L}(x_i, \mathcal{S}(\hat{p_i}, \beta_i)) = - \mathcal{S}(\hat{p_i}, \beta_i) \cdot \log \mathcal{F}_m(x_i)
$$
In practice, each $\mathcal{S}(\hat{p_i}, \beta_i)$ is computed only once as a preprocessing step. Our method \emph{does not require hyperparameters}, which can be an advantage since most out-of-distribution datasets do not provide a development set for tuning the hyperparameters.

%% file: exp.tex
%\section{Experiments}
%In this section, we demonstrate the effectiveness of our confidence regularization method in three different NLU tasks: natural language inference, fact verification, and paraphrase identification. In each task, we use the existing \textit{challenge datasets} to evaluate models' out-of-distribution performance.

\section{Experimental Setup}
In this section, we describe the datasets, models, and training details used in our experiments.
%\subsection{Datasets}
\subsection{Natural Language Inference} We use the MNLI dataset \cite{Williams2018ABC} for training. The dataset consists of pairs of premise and hypothesis sentences along with their inference labels (i.e., {entailment}, {neutral}, and {contradiction}). MNLI has two in-distribution development and test sets, one that matches domains of the training data (MNLI-m), and one with mismatching domains (MNLI-mm). We consider two out-of-distribution datasets for NLI: HANS (Heuristic Analysis for NLI Systems) \cite{McCoy2019RightFT} and MNLI-hard test sets \cite{Gururangan2018AnnotationAI}.

\paragraph{HANS} The dataset is constructed based on the finding that the word overlapping between premise and hypothesis in NLI datasets is strongly correlated with the \textit{entailment} label. HANS consists of examples in which such correlation does not exist, i.e., hypotheses are \textit{not entailed} by their word-overlapping premises. HANS is split into three test cases: (a) {Lexical overlap} (e.g., ``\textit{{The doctor} was {paid} by {the actor}}'' $\nRightarrow$ ``\textit{The doctor paid the actor}''), (b) {Subsequence} (e.g., ``\textit{The doctor near {the actor danced}}'' $\nRightarrow$ ``\textit{The actor danced}''), and (c) {Constituent} (e.g., ``\textit{If {the artist slept}, the actor ran}'' $\nRightarrow$ ``\textit{The artist slept}''). Each category contains both entailment and non-entailment examples. 
%Models that largely rely on the lexical-overlap bias will perform poorly on the three categories.

\paragraph{MNLI-hard} Hypothesis sentences in NLI datasets often contain words that are highly indicative of target labels \cite{Gururangan2018AnnotationAI, Poliak2018HypothesisOB}. It allows a simple model that predicts based on the hypothesis-only input to perform much better than the random baseline. \citet{Gururangan2018AnnotationAI} presents a ``hard'' split of the MNLI test sets, in which examples cannot be predicted correctly by the simple hypothesis-only model. 
%They demonstrate that models' performance on the hard split of the test set is substantially lower.

\subsection{Fact Verification} For this task, we use the training dataset provided by the FEVER challenge \cite{Thorne2018TheFE}. The task concerns about assessing the validity of a claim sentence in the context of a given evidence sentence, which can be labeled as either \textit{support}, \textit{refutes}, and \textit{not enough information}. 
% We train models on the pre-processed claim-evidence pairs provided by \citet{schuster2019towards}. 
We use the Fever-Symmetric dataset \cite{schuster2019towards} for the out-of-distribution evaluation.

\paragraph{Fever-Symmetric} \citet{schuster2019towards} introduce this dataset to demonstrate that FEVER models mostly rely on the claim-only bias, i.e., the occurrence of words and phrases in the claim that are biased toward certain labels. The dataset is manually constructed such that relying on cues of the claim can lead to incorrect predictions. We evaluate the models on the two versions (version 1 and 2) of their test sets.\footnote{\url{https://github.com/TalSchuster/FeverSymmetric}}

\subsection{Paraphrase Identification} We use the Quora Question Pairs (QQP) dataset for training. QQP consists of pairs of questions which are labeled as \textit{duplicate} if they are paraphrased, and \textit{non-duplicate} otherwise. We evaluate the out-of-distribution performance of QQP models on the QQP subset of PAWS (Paraphrase Adversaries from Word Scrambling) \cite{Zhang2019PAWSPA}.

\paragraph{PAWS} The QQP subset of PAWS consists of question pairs that are highly overlapping in words. The majority of these question pairs are labeled as non-duplicate. Models trained on QQP are shown to perform worse than the random baseline on this dataset. This partly indicates that models largely rely on lexical-overlap features to perform well on QQP. We report models' performance on the duplicate and non-duplicate examples separately. 

\begin{table*}[htb]
    \centering
    \small
    \setlength{\tabcolsep}{3pt}
    % \resizebox{\textwidth}{!}{%
    \begin{tabular}{r | c c | c c || c c c | c | c c}
        \toprule
         \multirow{2}{*}{Method} & \multicolumn{2}{c|}{MNLI-m} & \multicolumn{2}{c||}{MNLI-mm} & \multicolumn{4}{c|}{HANS} & \multicolumn{2}{c}{Hard subset} \\
        & dev & test & dev & test & lex. & subseq. & const. & \textit{avg.} & MNLI-m & MNLI-mm\\
        \midrule
        BERT-base & 84.3 \tiny{$\pm$ 0.3} &  84.6 &  84.7 \tiny{$\pm$ 0.1} & 83.3 &   72.4 &  52.7 & 57.9 &  61.1 \tiny{$\pm$ 1.1} &  76.8 &  75.9\\
        \midrule
        \midrule
        
        Learned-mixin\textsubscript{ hans} &  84.0 \tiny{$\pm$ 0.2} & 84.3 &  84.4  \tiny{$\pm$ 0.3} & 83.3 &  \textbf{77.5} &  54.1 &  63.2 &  64.9 \tiny{$\pm$ 2.4} & - & -\\ % seed 3333
        
        Product-of-expert\textsubscript{ hans} &  82.8 \tiny{$\pm$ 0.2} &  83.0 &   83.1 \tiny{$\pm$ 0.3} &  82.1 &  72.9 &  65.3 &  \textbf{69.6} &  \textbf{69.2} \tiny{$\pm$ 2.6} & - & -\\ % seed 5555 sub
        
        \midrule
        \textbf{Regularized-conf\textsubscript{ hans}} &  84.3 \tiny{$\pm$ 0.1} &  \textbf{84.7} &  84.8 \tiny{$\pm$ 0.2} &  83.4 & 73.3 & \textbf{66.5}  &  67.2 &  \textbf{69.1} \tiny{$\pm$ 1.2} & - & -\\
        \midrule
        
        Learned-mixin\textsubscript{ hypo} & 80.5 \tiny{$\pm$ 0.4} & 79.5 &  81.2 \tiny{$\pm$ 0.4} &  80.4 & - & - & - & - &  79.2 &  78.2 \\ % seed 222
        
        Product-of-expert\textsubscript{ hypo} &  83.5 \tiny{$\pm$ 0.4} &  82.8 &  83.8 \tiny{$\pm$ 0.2} & 84.1 & - & - & - & - & \textbf{79.8} & \textbf{78.7}\\
        
        \midrule
        \textbf{Regularized-conf\textsubscript{ hypo}} &  \textbf{84.6} \tiny{$\pm$ 0.2} &  84.1 &   \textbf{85.0} \tiny{$\pm$ 0.2} &  \textbf{84.2} & - & - & - & - & 78.3 &
        77.3\\
        % Regularized-conf\textsubscript{ mix} & 84.11 & - & 84.91 & -\\
        \bottomrule
    \end{tabular}%}
    \caption{The in-distribution accuracy (in percentage point) of the NLI models along with their accuracy on out-of-distribution test sets: HANS and MNLI hard subsets. Models are only evaluated against their targeted out-of-distribution dataset.}
    \label{tab:results_mnli}
\end{table*}

\subsection{Models}
\paragraph{Baseline Model} We apply all of the debiasing methods across our experiments on the BERT base model \cite{devlin2018bert}, which has shown impressive in-distribution performance on the three tasks. In our method, BERT base is used for both $\mathcal{F}_t$ and $\mathcal{F}_m$. We follow the standard setup for sentence pair classification tasks, in which the two sentences are concatenated into a single input  and the special token \texttt{[CLF]} is used for classification. 
% explain here why BERT
% mention that we use the same configuration across tasks
% our method does not use hyperparameter, so no search

\paragraph{Biased Model ($\mathcal{F}_b$)} We consider the biased features of each of the examined out-of-distribution datasets to train the biased models. For HANS and PAWS, we use hand-crafted features that indicate how words are shared between the two input sentences. Following \citet{Clark2019DontTT}, these features include the percentage of hypothesis words that also occur in the premise and the average of cosine distances between word embedding in the premise and hypothesis.\footnote{We include the detailed description in the appendix.} We then train a simple nonlinear classifier using these features. We refer to this biased model as the \textit{hans} model.

For MNLI-hard and Fever-Symmetric, we train a biased model on only hypothesis sentences and claim sentences for MNLI and FEVER, respectively. The biased model is a nonlinear classifier trained on top of the vector representation of the input sentence. We obtain this vector representation by max-pooling word embeddings into a single vector for FEVER, and by learning an LSTM-based sentence encoder for MNLI.

\paragraph{State-of-the-art Debiasing Models} We compare our method against existing state-of-the-art debiasing methods: \emph{product-of-expert} \cite{He2019UnlearnDB, Mahabadi2019SimpleBE} and its variant \emph{learned-mixin} \cite{Clark2019DontTT}. \emph{product-of-expert} ensembles the prediction of the main model ($p_i$) with the prediction of the biased model ($b_i$) using $p_i' = softmax(\log p_i + \log b_i)$, where $p_i'$ is the ensembled output distribution. This ensembling enables the main model to focus on learning from examples that are not predicted well by the biased model. \emph{Learned-mixin} improves this method by parameterizing the ensembling operation to let the model learn when to incorporate or ignore the output of the biased model for the ensembled prediction.

On FEVER, we also compare our method against the \emph{example-reweighting} method by \citet{schuster2019towards}. They compute the importance weight of each example based on the correlation of the n-grams within the claim sentences with the target labels. These weights are then used to compute the loss of each training batch.

\paragraph{Training Details} As observed by \citet{mccoy2019berts}, models can show high variance in their out-of-distribution performance. Therefore, we run each experiment five times and report both average and standard deviation of the scores.\footnote{Due to the limited number of possible submissions, we report the MNLI test scores only from a model that holds the median out-of-distribution performance.} We also use training configurations that are known to work well for each task.\footnote{We set a learning rate of $5e^{-5}$ for MNLI and $2e^{-5}$ for FEVER and QQP.}
%, and we do not perform early stopping. 
For each experiment, we train our \emph{confidence regularization} method as well as \emph{product-of-expert} and \emph{learned-mixin} using the {same} biased-model. Since the challenge datasets often do not provide a development set, we could not tune the hyperparameter of learned-mixin. We, therefore, use their default weight for the entropy penalty term.\footnote{E.g., $w=0.03$ for training on MNLI.}

\section{Results}
The results for the tasks of NLI, fact verification, and paraphrase identification are reported in Table~\ref{tab:results_mnli}, Table~\ref{tab:fever}, and Table~\ref{tab:qqp_paws}, respectively. 
%We use the mentioned \textit{challenge datasets} to evaluate models' out-of-distribution performance.
\subsection{In-distribution Performance} 
 The results on the original development and test sets of each task represent the in-distribution performance.
Since we examine two types of biases in NLI, we have two debiased NLI models, i.e., \emph{Regularized-conf}\textsubscript{hans} and \emph{Regularized-conf}\textsubscript{hypo} which are trained for debiasing HANS and hypothesis-only biases, respectively. 

We make the following observations from the results: (1) Our method outperforms \emph{product-of-expert} and \emph{learned-mixin} when evaluated on the corresponding in-distribution data of all the three tasks; (2) \emph{Product-of-expert} and \emph{learned-mixin} drop the original BERT baseline accuracy on most of the in-distribution experiments; (3) Regardless of the type of bias, our method preserves the in-distribution performance. However, it is not the case for the other two methods, e.g., \emph{learned-mixin} only results in a mild decrease in the accuracy when it is debiased for HANS, but suffers from substantial drop when it is used to address the hypothesis-only bias; (4) Our method results in a slight in-distribution improvement in some cases, e.g., on FEVER, it gains 0.6pp over BERT baseline. The models produced by \emph{Regularized-conf}\textsubscript{ hans} also gain 0.1 points to both MNLI-m and MNLI-mm test sets; (5) All methods, including ours decrease the in-distribution performance on QQP, particularly on its duplicate examples subset. We will discuss this performance drop in Section~\ref{sec:qqp_paws}.

\begin{table}
    \centering
    \setlength{\tabcolsep}{3pt}
    \resizebox{\columnwidth}{!}{%
    \begin{tabular}{r c || c c}
        \toprule
         Method & FEVER\textsubscript{ dev} & Symm.\textsubscript{ v1} & Symm.\textsubscript{ v2}\\
        \midrule
        BERT-base &  85.8 \footnotesize{$\pm$ 0.1} &  57.9 \footnotesize{$\pm$ 1.1} & 64.4 \footnotesize{$\pm$ 0.6}\\
        \midrule
        \midrule
        
        Learned-mixin\textsubscript{ claim} &  83.1 \footnotesize{$\pm$ 0.7} &  60.4 \footnotesize{$\pm$ 2.4} & 64.9 \footnotesize{$\pm$ 1.6}\\
        
        Product-of-expert\textsubscript{ claim} &  83.3 \footnotesize{$\pm$ 0.3} &  61.7 \footnotesize{$\pm$ 1.5} &  65.5 \footnotesize{$\pm$ 0.7}\\
        \midrule
        
        Reweighting\textsubscript{ bigrams} & 85.5 \footnotesize{$\pm$ 0.3} &  \textbf{61.7} \footnotesize{$\pm$ 1.1} & \textbf{66.5} \footnotesize{$\pm$ 1.3}\\
        \midrule
        
        \textbf{Regularized-conf\textsubscript{ claim}} &  \textbf{86.4} \footnotesize{$\pm$ 0.2} & 60.5 \footnotesize{$\pm$ 0.4} & 66.2 \footnotesize{$\pm$ 0.6}\\
        \bottomrule
    \end{tabular}}
    \caption{Accuracy on the FEVER dataset and the corresponding challenge datasets.}
    \label{tab:fever}
\end{table}

\subsection{Out-of-distribution Performance} 
The rightmost columns of each table report the evaluation results on the out-of-distribution datasets for each task. 
Based on our out-of-distribution evaluations, we observe that: (1) Our method minimizes the trade-off between the in-distribution and out-of-distribution performance compared to the other methods. For example, on HANS, \emph{learned-mixin} maintains the in-distribution performance but only improves the average HANS accuracy from 61.1\% to 64.9\%. \emph{product-of-expert} gains 7 points improvement over the BERT baseline while reducing the MNLI-m test accuracy by 1.6 points. On the other hand, our method achieves the competitive 7 points gain without dropping the in-distribution performance;
(2) The performance trade-off is stronger on some datasets. On PAWS, the two compared methods improve the accuracy on the \emph{non-duplicate} subset while reducing models' ability to detect the \emph{duplicate} examples. Our method, on the other hand, finds a balance point, in which the non-duplicate accuracy can no longer be improved without reducing the duplicate accuracy;
(3) depending on the use of hyperparameters, \emph{learned-mixin} can make a lower out-of-distribution improvement compared to ours, even after substantially degrading in-distribution performance, e.g., on FEVER-symmetric\textsubscript{v2}, it only gains 0.5 points while dropping 3 points on the FEVER development set. 

\begin{table}
    \centering
    \setlength{\tabcolsep}{3pt}
    \resizebox{\columnwidth}{!}{%
    \begin{tabular}{r c c || c c}
        \toprule
         \multirow{2}{*}{Method} & \multicolumn{2}{c||}{QQP dev} & \multicolumn{2}{c}{PAWS test} \\
        & dupl & \textbf{$\neg$}dupl & dupl & \textbf{$\neg$}dupl \\
        \midrule
        BERT-base & 88.4 \tiny{$\pm$ 0.3} & 92.5 \tiny{$\pm$ 0.3} &  96.9 \tiny{$\pm$ 0.3} &  9.8 \tiny{$\pm$ 0.4}\\
        \midrule
        \midrule
        LMixin\textsubscript{ hans} &  77.5 \tiny{$\pm$ 0.7} &  91.9 \tiny{$\pm$ 0.2} &  69.7 \tiny{$\pm$ 4.3} &  \textbf{51.7} \tiny{$\pm$ 4.3}\\
        Prod-exp\textsubscript{ hans} &  80.8 \tiny{$\pm$ 0.2} &  \textbf{93.5} \tiny{$\pm$ 0.1} & 71.0 \tiny{$\pm$ 2.3} & 49.9 \tiny{$\pm$ 2.3}\\
        \midrule
        \textbf{Reg-conf\textsubscript{ hans}} & \textbf{85.0} \tiny{$\pm$ 0.7} &  91.5 \tiny{$\pm$ 0.4} &  \textbf{91.0} \tiny{$\pm$ 1.8} &  19.8 \tiny{$\pm$ 1.3}\\
        % Regularized-conf\textsubscript{ mix} & 77.02 & 77.90 & 78.51 & 77.30\\
        \bottomrule
    \end{tabular}}
    \caption{Results of the evaluation on the QQP task.}
    \label{tab:qqp_paws}
\end{table}
% add some point about the need of appropriate hyperparameter here? 

%% file: discussion.tex
\section{Discussions and Analysis}
% \begin{figure*}
%     \centering
%     \includegraphics[width=13cm]{figures/confidence_dist.pdf}
%     \caption{The distribution of the output probability argmax from each model. The blue bars indicate the fraction of the prediction within each bin that are correct. All predictions are taken by evaluating the models on non-entailment subsequence subset of HANS.}
%     \label{fig:confidence_dist}
%     \vspace{-4mm}
% \end{figure*}
% \paragraph{Comparison to the state-of-the-arts}
\begin{figure*}
    \centering
    \begin{subfigure}[b]{0.9\textwidth}
       \includegraphics[width=1\linewidth]{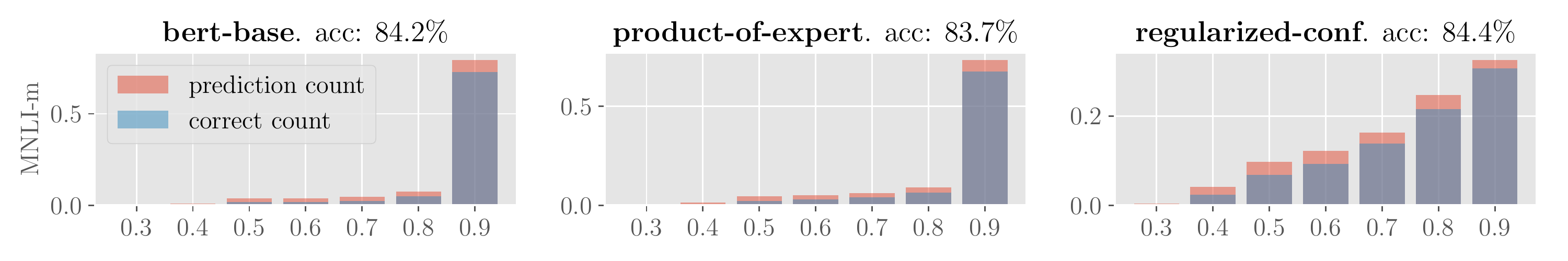}
       \caption{}
       \label{fig:conf1} 
    \end{subfigure}
    \begin{subfigure}[b]{0.9\textwidth}
       \includegraphics[width=1\linewidth]{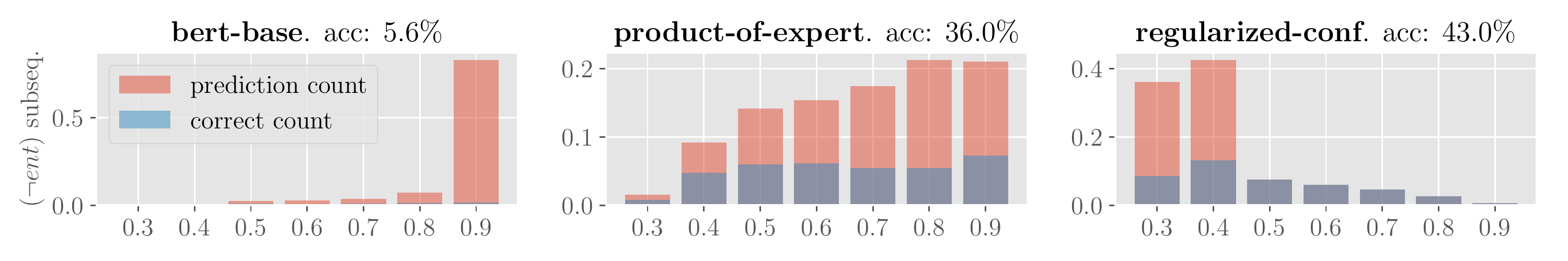}
       \caption{}
       \label{fig:conf2} 
    \end{subfigure}
    \caption{Distribution of models' confidence on their predicted labels. The blue areas indicate the fraction of each bin that are correct. (a) Distribution on MNLI-m dev by models trained using hypothesis-only biased model. (b) Distribution on non-entailment subsequence subset of HANS by models trained using \textit{hans} biased-model.}
    \label{fig:confidence_dist}
\end{figure*}

\paragraph{Ablation studies}
In this section, we show that the resulting improvements from our method come from the combination of both self-distillation and our scaling mechanism.
We perform ablation studies to examine the impact of each of the components including (1) \emph{self-distillation}: we train a model using the standard self-distillation without bias-weighted scaling, and (2) \emph{example-reweighting}: we train a model with the standard cross-entropy loss with an example reweighting method to adjust the importance of individual examples to the loss. The weight of each example is obtained from the (scaled) probability that is assigned by the teacher model to the ground truth label.\footnote{Details of the ablation experiments are included in the supplementary materials.} The aim of the second setting is to exclude the effect of self-distillation while keeping the effect of our scaling mechanism.

Table \ref{tab:treatments} presents the results of these experiments on MNLI and HANS. We observe that each component individually still gains substantial improvements on HANS over the baseline, albeit not as strong as the full method. The results from the \emph{self-distillation} suggest that the improvement from our method partly comes from the regularization effect of the distillation objective \cite{Clark2019BAMBM, Furlanello2018BornAgainNN}. 
In the \emph{example-reweighting} experiment, we exclude the effect of all the scaled teacher's output except for the probability assigned to the ground truth label. Compared to \emph{self-distillation}, the proposed \emph{example-reweighting} has a higher impact on improving the performance in both in-distribution and out-of-distribution evaluations.
However, both components are necessary for the overall improvements. 

\begin{table}
    \centering
    \small
    \setlength{\tabcolsep}{3pt}
    % \resizebox{0.5\columnwidth}{!}{%
    \begin{tabular}{l rr}
        \toprule
         Method & MNLI & HANS \\
        \midrule
        BERT-base & 84.3 & 61.1\\
        \midrule
        Full method & 84.3 & 69.1\\
        \midrule
        self-distillation & 84.6 & 64.4\\
        example-reweighting & 84.7 & 65.3\\
        %permuted & 83.9 & 64.5\\
        \bottomrule
    \end{tabular}%}
    \caption{Results of the ablation experiments. The MNLI column refers to the MNLI-m dev set.}
    \label{tab:treatments}
    % \vspace{-4mm}
\end{table}

\begin{table}
    \centering
    \setlength{\tabcolsep}{1pt}
    \resizebox{\columnwidth}{!}{%
    \begin{tabular}{l P{2cm} | P{2cm} | P{2cm} | P{2cm}}
        \toprule
           & BERT-baseline & product-of-expert & learned-mixin & \textbf{conf-reg (our)}\\
        \midrule
        MNLI-m & 9.0 & 7.7 & 9.9 & \textbf{5.4}\\
        MNLI-mm & 8.5 & 7.6 & 9.5 & \textbf{5.6}\\
        \bottomrule
    \end{tabular}}
    \caption{The calibration scores of models measured by ECE (lower is better).} 
    \label{tab:calibration}
    % \vspace{-4mm}
\end{table}
\paragraph{In-distribution performance drop of product-of-expert}
The difference between our method with \emph{product-of-expert} and its variants is the use of biased examples during training. \emph{Product-of-expert} in practice scales down the gradients on the biased training examples to allow the model to focus on learning from the harder examples \cite{He2019UnlearnDB}. As a result, models often receive little to no incentive to solve these examples throughout the training, which can effectively reduce the training data size. Our further examination on a \emph{product-of-expert} model (trained on MNLI for HANS) shows that its degradation of in-distribution performance largely comes from the aforementioned examples. Ensembling back the \textit{biased-model} to the main model can indeed bring the in-distribution accuracy back to the BERT baseline. However, this also leads to the original poor performance on HANS, which is counterproductive to the goal of improving the out-of-distribution generalization.
% which advantage

\paragraph{Impact on Models' Calibration}
We expect the training objective used in our method to discourage models from making overconfident predictions, i.e., assigning high probability to the predicted labels even when they are incorrect. We investigate the changes in models' behavior in terms of their confidence using the measure of \emph{calibration}, which quantifies how aligned the confidence of the predicted labels with their actual accuracy are \cite{Guo2017OnCO}. We compute the \emph{expected calibration error} (ECE) \cite{Naeini2015ObtainingWC} as a scalar summary statistic of calibration. Results in Table~\ref{tab:calibration} show that our method improves model's calibration on MNLI-m and MNLI-mm dev sets, with the reduction of ECE ranging from 3.0 to 3.6. The histograms in figure \ref{fig:confidence_dist} show the distribution of models' confidences in their predictions. Figure~\ref{fig:conf1} demonstrates that the prediction confidences of our resulting model on MNLI-m are more smoothly distributed. In figure \ref{fig:conf2}, we observe that our debiased model predicts examples that contain lexical overlap features with lower confidence, and when the confidence is higher, the prediction is more likely to be correct.

\paragraph{Impact of biased examples ratio}\label{sec:qqp_paws} To investigate the slight in-distribution drop by our method in QQP (Table~\ref{tab:qqp_paws}), we examine the ratio of biased examples in the QQP training data by evaluating the performance of the biased model on the dataset. 
We find that almost 80\% of the training examples can be solved using the lexical overlap features alone, which indicates a severe lexical overlap bias in QQP.\footnote{The random baseline is 50\% for QQP.}
Moreover, in 53\% of all examples, the biased model makes correct predictions with a very high confidence ($\beta_i > 0.8$). For comparison, the same biased model predicts only 12\% of the MNLI examples with confidence above 0.8 (more comparisons are shown in the supplementary material. As a result, there are not enough unbiased examples in QQP and 
the resulting soft target labels in this dataset are mostly close to a uniform distribution, which in turn may provide insufficient training signal to maximize the accuracy on the training distribution.

\begin{figure}
    \centering
    \includegraphics[width=1.\columnwidth]{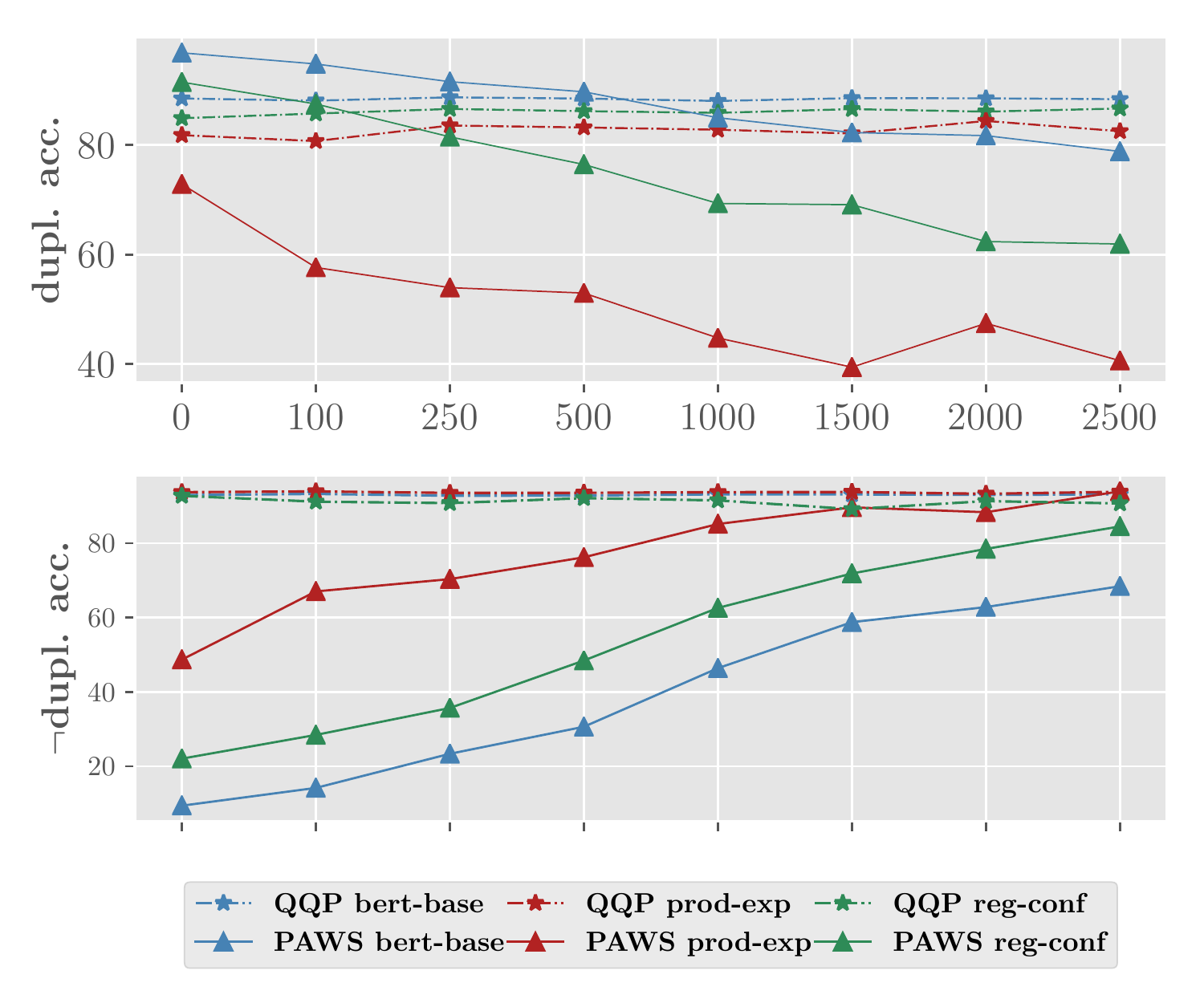}
    \caption{Results on the PAWS-augmented QQP dataset.}
    % The x-axis 0 correspond to the original QQP dataset, the rightmost point correspond to models trained on all of 2500 PAWS training split.}
    \label{fig:paws_augmented}
\end{figure}
\paragraph{Impact of adding bias-free examples} Finally, we investigate how changing the ratio of biased examples affects the behavior of debiasing methods. To this end, we split PAWS data into training and test sets.
The training set consists of 2500 examples, and we use the remaining 10K examples as a test set. We train the model on QQP that is gradually augmented with fractions of this PAWS training split and evaluate on a constant PAWS test set. Figure~\ref{fig:paws_augmented} shows the results of this experiment. 
When more PAWS examples are added to the training data, the accuracy of the BERT baseline gradually improves on the non-duplicate subset while its accuracy slowly drops on the duplicate subset. 
We observe that \emph{product-of-expert} exaggerates this effect: it reduces the duplicate accuracy up to 40\% to obtain the 93\% non-duplicate accuracy. We note that our method is the most effective when the entire 2500 PAWS examples are included in the training, obtaining the overall accuracy of 77.05\% compared to the 71.63\% from the baseline BERT.

%% file: appendix.tex
\begin{figure*}[!htb]
    \centering
    \begin{subfigure}[b]{0.3\linewidth}
       \includegraphics[width=1\linewidth]{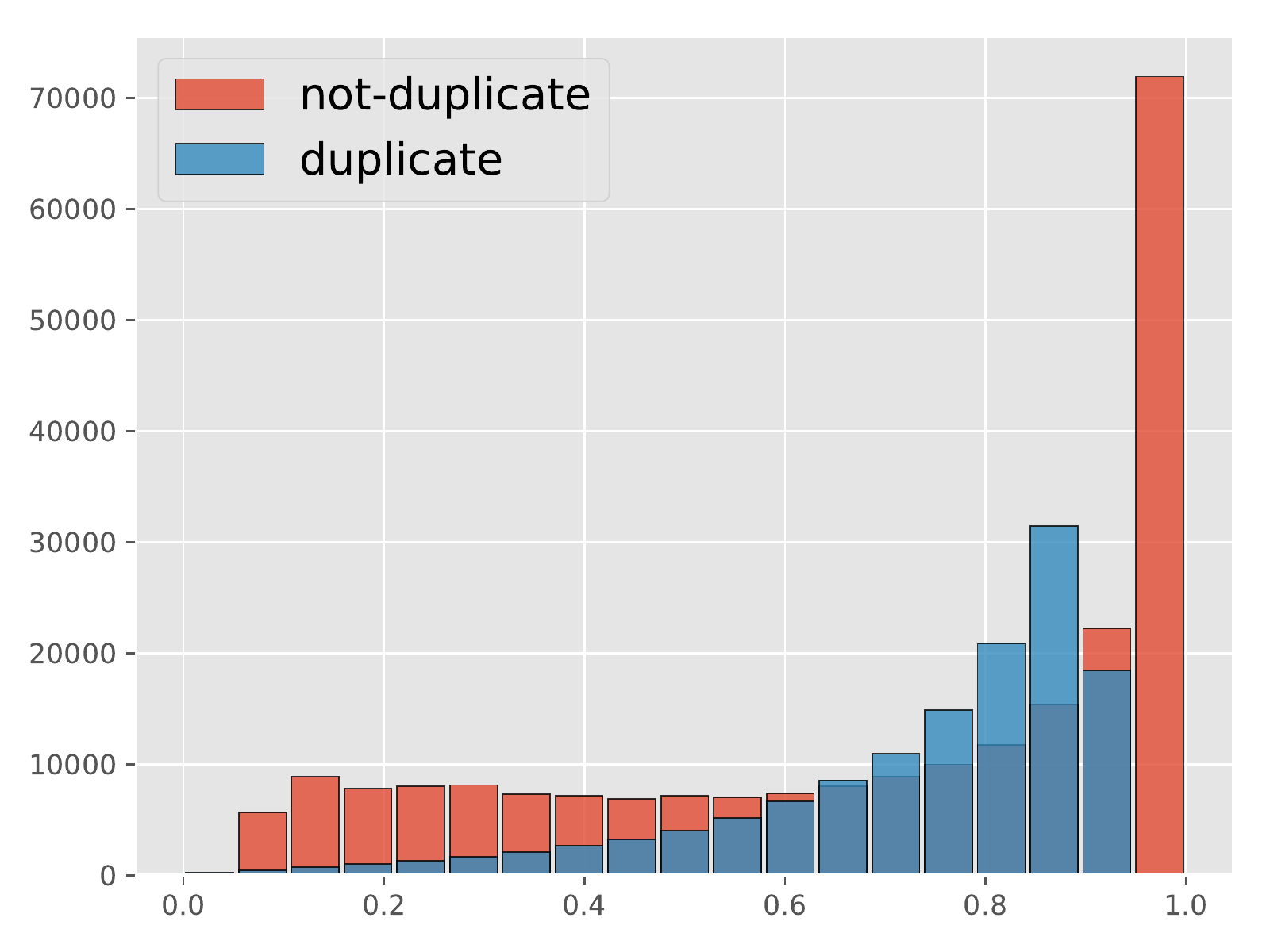}
       \caption{}
       \label{fig:beta_dist_1} 
    \end{subfigure}
    \begin{subfigure}[b]{0.3\linewidth}
       \includegraphics[width=1\linewidth]{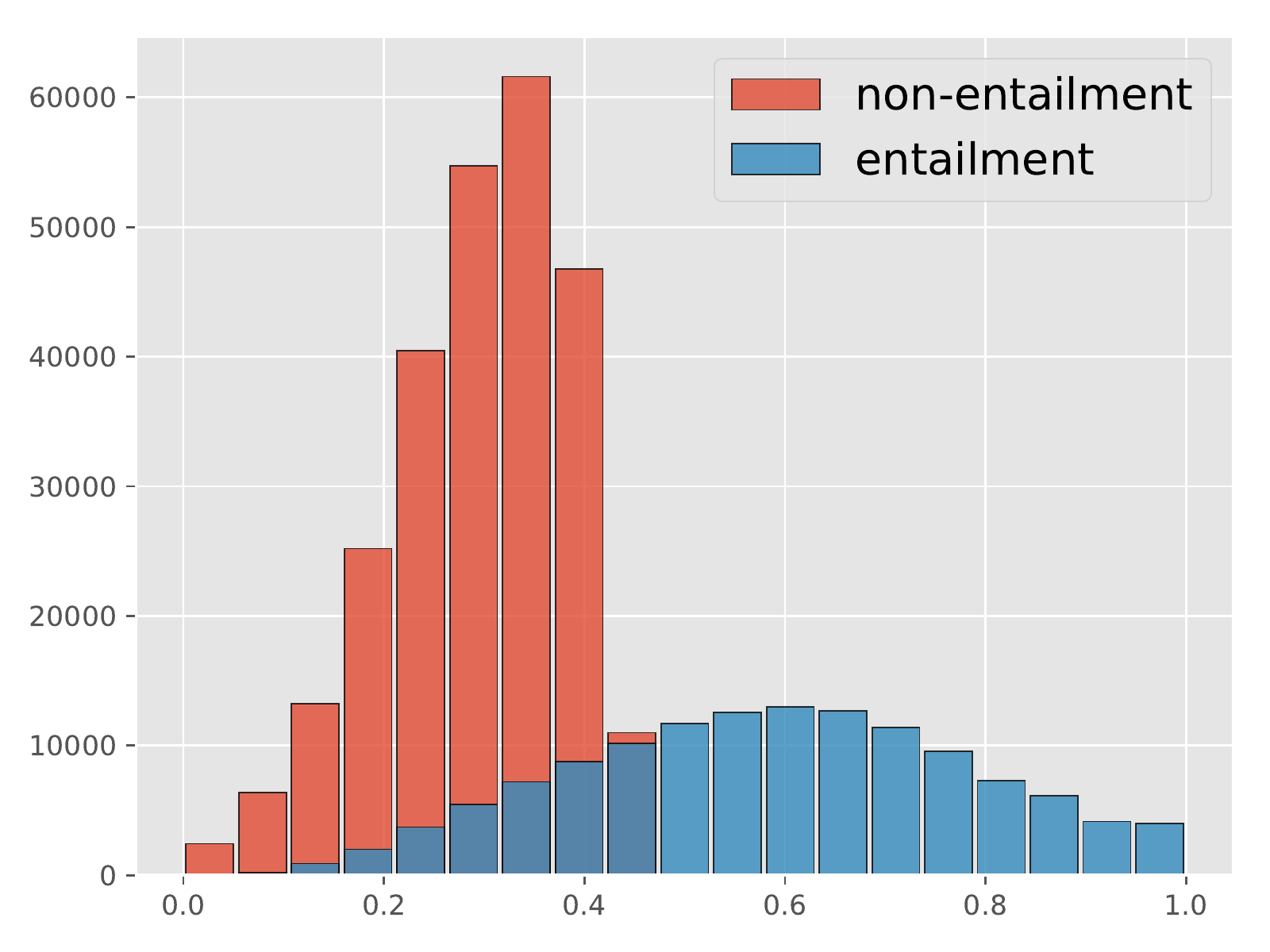}
       \caption{}
       \label{fig:beta_dist_2} 
    \end{subfigure}
    \begin{subfigure}[b]{0.3\linewidth}
       \includegraphics[width=1\linewidth]{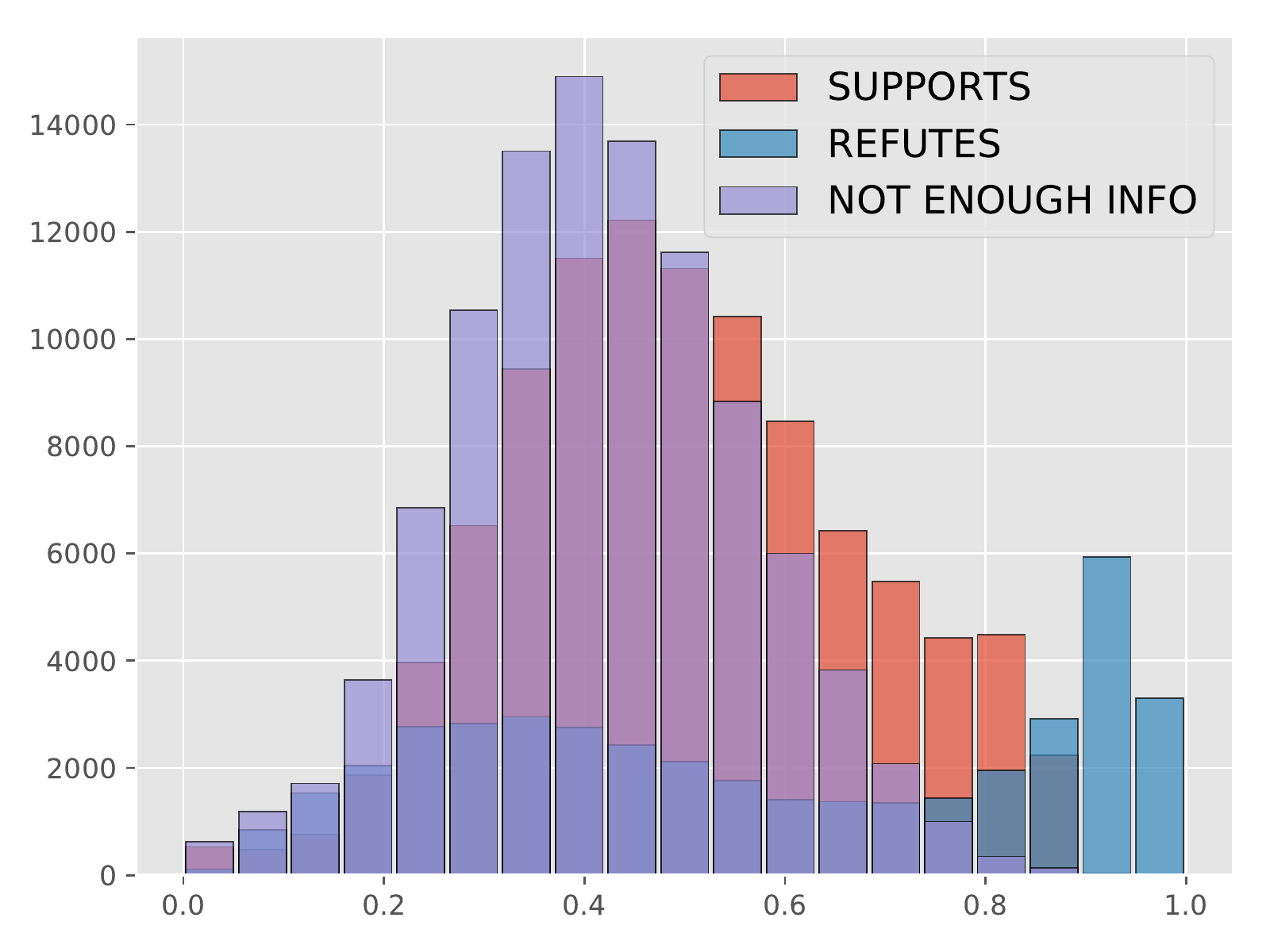}
       \caption{}
       \label{fig:beta_dist_3} 
    \end{subfigure}
    \caption{The distribution of biased model confidence on three training datasets of QQP, MNLI, and FEVER.}
    \label{fig:beta_dist}
\end{figure*}

\section{Ablation Details}
\label{sec:appendix_treatment} For the second setting of our ablation studies, we perform an example reweighting using the scaled probability of the teacher model $\mathcal{F}_t$ on the ground truth label. Specifically, the cross entropy loss assigned to each batch of size $m$ is computed by the following:
$$
    -\sum_{s=1}^b \frac{\hat{p_{s,c}}}{\sum_{u=1}^b \hat{p_{u,c}}} \cdot \log (p_{s,c})
$$
where we assume that $c$th label is the ground truth label. The probability assigned to the correct label by the teacher model is then denoted as $\hat{p_{s,c}}$. The currect predicted probability of the main model is denoted as $p_{s,c}$.

\section{Bias Weights Distribution}
\label{sec:appendix_w_dist} Figure \ref{fig:beta_dist} shows the performance of biased models on QQP, MNLI, and FEVER. For QQP and MNLI we show the results of biased model trained using lexical overlap features. For FEVER, the biased model is trained with claim-only partial input. We show that on PAWS (figure \ref{fig:beta_dist_1}), a large portion of examples can be predicted with a very high confidence by the biased model.

\section{HANS Biased Model}
We use the hand-crafted HANS-based features proposed by \citet{Clark2019DontTT}. These features include: (1) whether all words in the hypothesis exist in the premise; (2) whether the hypothesis is a contiguous subsequence of the premise; (3) the fraction of hypothesis words that exist in the premise; (4) the average and the max of cosine distances between word vectors in the premise and the hypothesis.

% \section{Computing Calibration}
% \label{sec:appendix_calibration} Following \citet{Guo2017OnCO}, the expected calibration error (ECE) can be computed by the following. 

% \section{TODOs}
% \label{sec:supplemental}
% \begin{enumerate}
%     \item Abstract and conclusion
%     \item Motivate the use of knowledge distillation [x]
%     \item Mention about the overconfident problem in the intro [x]
%     \item Make more concise description of our method in the related work [x]
% \end{enumerate}

%% file: acl2020.bbl
\begin{thebibliography}{30}
\expandafter\ifx\csname natexlab\endcsname\relax\def\natexlab#1{#1}\fi

\bibitem[{Agrawal et~al.(2016)Agrawal, Batra, and
  Parikh}]{agrawal-etal-2016-analyzing}
Aishwarya Agrawal, Dhruv Batra, and Devi Parikh. 2016.
\newblock \href {https://doi.org/10.18653/v1/D16-1203} {Analyzing the behavior
  of visual question answering models}.
\newblock In \emph{Proceedings of the 2016 Conference on Empirical Methods in
  Natural Language Processing}, pages 1955--1960, Austin, Texas. Association
  for Computational Linguistics.

\bibitem[{Belinkov et~al.(2019)Belinkov, Poliak, Shieber, Durme, and
  Rush}]{belinkovAdv2019}
Yonatan Belinkov, Adam Poliak, Stuart~M. Shieber, Benjamin~Van Durme, and
  Alexander~M. Rush. 2019.
\newblock \href {https://doi.org/10.18653/v1/s19-1028} {On adversarial removal
  of hypothesis-only bias in natural language inference}.
\newblock In \emph{Proceedings of the Eighth Joint Conference on Lexical and
  Computational Semantics, *SEM@NAACL-HLT 2019, Minneapolis, MN, USA, June 6-7,
  2019}, pages 256--262. Association for Computational Linguistics.

\bibitem[{Bender and Koller(2020)}]{bender2020climbing}
Emily Bender and Alexander Koller. 2020.
\newblock Climbing towards {NLU}: On meaning, form, and understanding in the
  age of data.
\newblock In \emph{Proceedings of the 58th Annual Meeting of the Association
  for Computational Linguistics}, page \textit{to appear}, virtual conference.
  Association for Computational Linguistics.

\bibitem[{Clark et~al.(2019{\natexlab{a}})Clark, Yatskar, and
  Zettlemoyer}]{Clark2019DontTT}
Christopher Clark, Mark Yatskar, and Luke Zettlemoyer. 2019{\natexlab{a}}.
\newblock \href {https://doi.org/10.18653/v1/D19-1418} {Don{'}t take the easy
  way out: Ensemble based methods for avoiding known dataset biases}.
\newblock In \emph{Proceedings of the 2019 Conference on Empirical Methods in
  Natural Language Processing and the 9th International Joint Conference on
  Natural Language Processing (EMNLP-IJCNLP)}, pages 4067--4080, Hong Kong,
  China. Association for Computational Linguistics.

\bibitem[{Clark et~al.(2019{\natexlab{b}})Clark, Luong, Khandelwal, Manning,
  and Le}]{Clark2019BAMBM}
Kevin Clark, Minh-Thang Luong, Urvashi Khandelwal, Christopher~D. Manning, and
  Quoc~V. Le. 2019{\natexlab{b}}.
\newblock \href {https://doi.org/10.18653/v1/P19-1595} {{BAM}! born-again
  multi-task networks for natural language understanding}.
\newblock In \emph{Proceedings of the 57th Annual Meeting of the Association
  for Computational Linguistics}, pages 5931--5937, Florence, Italy.
  Association for Computational Linguistics.

\bibitem[{Devlin et~al.(2019)Devlin, Chang, Lee, and
  Toutanova}]{devlin2018bert}
Jacob Devlin, Ming-Wei Chang, Kenton Lee, and Kristina Toutanova. 2019.
\newblock \href {https://doi.org/10.18653/v1/N19-1423} {{BERT}: Pre-training of
  deep bidirectional transformers for language understanding}.
\newblock In \emph{Proceedings of the 2019 Conference of the North {A}merican
  Chapter of the Association for Computational Linguistics: Human Language
  Technologies, Volume 1 (Long and Short Papers)}, pages 4171--4186,
  Minneapolis, Minnesota. Association for Computational Linguistics.

\bibitem[{Falke et~al.(2019)Falke, Ribeiro, Utama, Dagan, and
  Gurevych}]{falke-etal-2019-ranking}
Tobias Falke, Leonardo F.~R. Ribeiro, Prasetya~Ajie Utama, Ido Dagan, and Iryna
  Gurevych. 2019.
\newblock \href {https://doi.org/10.18653/v1/P19-1213} {Ranking generated
  summaries by correctness: An interesting but challenging application for
  natural language inference}.
\newblock In \emph{Proceedings of the 57th Annual Meeting of the Association
  for Computational Linguistics}, pages 2214--2220, Florence, Italy.
  Association for Computational Linguistics.

\bibitem[{Feng et~al.(2018)Feng, Wallace, Grissom~II, Iyyer, Rodriguez, and
  Boyd-Graber}]{Feng2018PathologiesON}
Shi Feng, Eric Wallace, Alvin Grissom~II, Mohit Iyyer, Pedro Rodriguez, and
  Jordan Boyd-Graber. 2018.
\newblock \href {https://doi.org/10.18653/v1/D18-1407} {Pathologies of neural
  models make interpretations difficult}.
\newblock In \emph{Proceedings of the 2018 Conference on Empirical Methods in
  Natural Language Processing}, pages 3719--3728, Brussels, Belgium.
  Association for Computational Linguistics.

\bibitem[{Furlanello et~al.(2018)Furlanello, Lipton, Tschannen, Itti, and
  Anandkumar}]{Furlanello2018BornAgainNN}
Tommaso Furlanello, Zachary~Chase Lipton, Michael Tschannen, Laurent Itti, and
  Anima Anandkumar. 2018.
\newblock \href {http://proceedings.mlr.press/v80/furlanello18a.html}
  {Born-again neural networks}.
\newblock In \emph{Proceedings of the 35th International Conference on Machine
  Learning, {ICML} 2018, Stockholmsm{\"{a}}ssan, Stockholm, Sweden, July 10-15,
  2018}, volume~80 of \emph{Proceedings of Machine Learning Research}, pages
  1602--1611. {PMLR}.

\bibitem[{Guo et~al.(2017)Guo, Pleiss, Sun, and Weinberger}]{Guo2017OnCO}
Chuan Guo, Geoff Pleiss, Yu~Sun, and Kilian~Q. Weinberger. 2017.
\newblock \href {http://proceedings.mlr.press/v70/guo17a.html} {On calibration
  of modern neural networks}.
\newblock In \emph{Proceedings of the 34th International Conference on Machine
  Learning, {ICML} 2017, Sydney, NSW, Australia, 6-11 August 2017}, volume~70
  of \emph{Proceedings of Machine Learning Research}, pages 1321--1330. {PMLR}.

\bibitem[{Gururangan et~al.(2018)Gururangan, Swayamdipta, Levy, Schwartz,
  Bowman, and Smith}]{Gururangan2018AnnotationAI}
Suchin Gururangan, Swabha Swayamdipta, Omer Levy, Roy Schwartz, Samuel Bowman,
  and Noah~A. Smith. 2018.
\newblock \href {https://doi.org/10.18653/v1/N18-2017} {Annotation artifacts in
  natural language inference data}.
\newblock In \emph{Proceedings of the 2018 Conference of the North {A}merican
  Chapter of the Association for Computational Linguistics: Human Language
  Technologies, Volume 2 (Short Papers)}, pages 107--112, New Orleans,
  Louisiana. Association for Computational Linguistics.

\bibitem[{He et~al.(2019)He, Zha, and Wang}]{He2019UnlearnDB}
He~He, Sheng Zha, and Haohan Wang. 2019.
\newblock \href {https://doi.org/10.18653/v1/D19-6115} {Unlearn dataset bias in
  natural language inference by fitting the residual}.
\newblock In \emph{Proceedings of the 2nd Workshop on Deep Learning Approaches
  for Low-Resource NLP, DeepLo@EMNLP-IJCNLP 2019, Hong Kong, China, November 3,
  2019}, pages 132--142. Association for Computational Linguistics.

\bibitem[{Hinton et~al.(2015)Hinton, Vinyals, and
  Dean}]{Hinton2015DistillingTK}
Geoffrey~E. Hinton, Oriol Vinyals, and Jeffrey Dean. 2015.
\newblock \href {http://arxiv.org/abs/1503.02531} {Distilling the knowledge in
  a neural network}.
\newblock \emph{CoRR}, abs/1503.02531.

\bibitem[{Kaushik et~al.(2020)Kaushik, Hovy, and Lipton}]{Kaushik2020Learning}
Divyansh Kaushik, Eduard Hovy, and Zachary Lipton. 2020.
\newblock \href {https://openreview.net/forum?id=Sklgs0NFvr} {Learning the
  difference that makes a difference with counterfactually-augmented data}.
\newblock In \emph{8th International Conference on Learning Representations,
  {ICLR} 2020, Virtual Conference, 26 April - 1 May, 2019}. OpenReview.net.

\bibitem[{Mahabadi and Henderson(2019)}]{Mahabadi2019SimpleBE}
Rabeeh~Karimi Mahabadi and James Henderson. 2019.
\newblock \href {http://arxiv.org/abs/1909.06321} {Simple but effective
  techniques to reduce biases}.
\newblock \emph{CoRR}, abs/1909.06321.

\bibitem[{McCoy et~al.(2019{\natexlab{a}})McCoy, Min, and
  Linzen}]{mccoy2019berts}
R~Thomas McCoy, Junghyun Min, and Tal Linzen. 2019{\natexlab{a}}.
\newblock Berts of a feather do not generalize together: Large variability in
  generalization across models with similar test set performance.
\newblock \emph{arXiv preprint arXiv:1911.02969}.

\bibitem[{McCoy et~al.(2019{\natexlab{b}})McCoy, Pavlick, and
  Linzen}]{McCoy2019RightFT}
Tom McCoy, Ellie Pavlick, and Tal Linzen. 2019{\natexlab{b}}.
\newblock \href {https://doi.org/10.18653/v1/P19-1334} {Right for the wrong
  reasons: Diagnosing syntactic heuristics in natural language inference}.
\newblock In \emph{Proceedings of the 57th Annual Meeting of the Association
  for Computational Linguistics}, pages 3428--3448, Florence, Italy.
  Association for Computational Linguistics.

\bibitem[{Naeini et~al.(2015)Naeini, Cooper, and
  Hauskrecht}]{Naeini2015ObtainingWC}
Mahdi~Pakdaman Naeini, Gregory~F. Cooper, and Milos Hauskrecht. 2015.
\newblock \href {http://www.aaai.org/ocs/index.php/AAAI/AAAI15/paper/view/9667}
  {Obtaining well calibrated probabilities using bayesian binning}.
\newblock In \emph{Proceedings of the Twenty-Ninth {AAAI} Conference on
  Artificial Intelligence, January 25-30, 2015, Austin, Texas, {USA}}, pages
  2901--2907. {AAAI} Press.

\bibitem[{Niven and Kao(2019)}]{niven2019probing}
Timothy Niven and Hung-Yu Kao. 2019.
\newblock \href {https://doi.org/10.18653/v1/P19-1459} {Probing neural network
  comprehension of natural language arguments}.
\newblock In \emph{Proceedings of the 57th Annual Meeting of the Association
  for Computational Linguistics}, pages 4658--4664, Florence, Italy.
  Association for Computational Linguistics.

\bibitem[{Papernot et~al.(2016)Papernot, McDaniel, Wu, Jha, and
  Swami}]{Papernot2015DistillationAA}
Nicolas Papernot, Patrick~D. McDaniel, Xi~Wu, Somesh Jha, and Ananthram Swami.
  2016.
\newblock \href {https://doi.org/10.1109/SP.2016.41} {Distillation as a defense
  to adversarial perturbations against deep neural networks}.
\newblock In \emph{{IEEE} Symposium on Security and Privacy, {SP} 2016, San
  Jose, CA, USA, May 22-26, 2016}, pages 582--597. {IEEE} Computer Society.

\bibitem[{Pereyra et~al.(2017)Pereyra, Tucker, Chorowski, Kaiser, and
  Hinton}]{Pereyra2017RegularizingNN}
Gabriel Pereyra, George Tucker, Jan Chorowski, Lukasz Kaiser, and Geoffrey~E.
  Hinton. 2017.
\newblock \href {https://openreview.net/forum?id=HyhbYrGYe} {Regularizing
  neural networks by penalizing confident output distributions}.
\newblock In \emph{5th International Conference on Learning Representations,
  {ICLR} 2017, Toulon, France, April 24-26, 2017, Workshop Track Proceedings}.
  OpenReview.net.

\bibitem[{Poliak et~al.(2018)Poliak, Naradowsky, Haldar, Rudinger, and
  Van~Durme}]{Poliak2018HypothesisOB}
Adam Poliak, Jason Naradowsky, Aparajita Haldar, Rachel Rudinger, and Benjamin
  Van~Durme. 2018.
\newblock \href {https://doi.org/10.18653/v1/S18-2023} {Hypothesis only
  baselines in natural language inference}.
\newblock In \emph{Proceedings of the Seventh Joint Conference on Lexical and
  Computational Semantics}, pages 180--191, New Orleans, Louisiana. Association
  for Computational Linguistics.

\bibitem[{Sakaguchi et~al.(2019)Sakaguchi, Bras, Bhagavatula, and
  Choi}]{winogrande2019}
Keisuke Sakaguchi, Ronan~Le Bras, Chandra Bhagavatula, and Yejin Choi. 2019.
\newblock \href {http://arxiv.org/abs/1907.10641} {{WINOGRANDE:} an adversarial
  winograd schema challenge at scale}.
\newblock \emph{CoRR}, abs/1907.10641.

\bibitem[{Schuster et~al.(2019)Schuster, Shah, Yeo, Roberto Filizzola~Ortiz,
  Santus, and Barzilay}]{schuster2019towards}
Tal Schuster, Darsh Shah, Yun Jie~Serene Yeo, Daniel Roberto Filizzola~Ortiz,
  Enrico Santus, and Regina Barzilay. 2019.
\newblock \href {https://doi.org/10.18653/v1/D19-1341} {Towards debiasing fact
  verification models}.
\newblock In \emph{Proceedings of the 2019 Conference on Empirical Methods in
  Natural Language Processing and the 9th International Joint Conference on
  Natural Language Processing (EMNLP-IJCNLP)}, pages 3417--3423, Hong Kong,
  China. Association for Computational Linguistics.

\bibitem[{Szegedy et~al.(2016)Szegedy, Vanhoucke, Ioffe, Shlens, and
  Wojna}]{Szegedy2015RethinkingTI}
Christian Szegedy, Vincent Vanhoucke, Sergey Ioffe, Jonathon Shlens, and
  Zbigniew Wojna. 2016.
\newblock \href {https://doi.org/10.1109/CVPR.2016.308} {Rethinking the
  inception architecture for computer vision}.
\newblock In \emph{2016 {IEEE} Conference on Computer Vision and Pattern
  Recognition, {CVPR} 2016, Las Vegas, NV, USA, June 27-30, 2016}, pages
  2818--2826. {IEEE} Computer Society.

\bibitem[{Thorne et~al.(2018)Thorne, Vlachos, Cocarascu, Christodoulopoulos,
  and Mittal}]{Thorne2018TheFE}
James Thorne, Andreas Vlachos, Oana Cocarascu, Christos Christodoulopoulos, and
  Arpit Mittal. 2018.
\newblock \href {https://doi.org/10.18653/v1/W18-5501} {The fact extraction and
  {VER}ification ({FEVER}) shared task}.
\newblock In \emph{Proceedings of the First Workshop on Fact Extraction and
  {VER}ification ({FEVER})}, pages 1--9, Brussels, Belgium. Association for
  Computational Linguistics.

\bibitem[{Wang et~al.(2018)Wang, Singh, Michael, Hill, Levy, and
  Bowman}]{wang2018glue}
Alex Wang, Amanpreet Singh, Julian Michael, Felix Hill, Omer Levy, and Samuel
  Bowman. 2018.
\newblock \href {https://doi.org/10.18653/v1/W18-5446} {{GLUE}: A multi-task
  benchmark and analysis platform for natural language understanding}.
\newblock In \emph{Proceedings of the 2018 {EMNLP} Workshop {B}lackbox{NLP}:
  Analyzing and Interpreting Neural Networks for {NLP}}, pages 353--355,
  Brussels, Belgium. Association for Computational Linguistics.

\bibitem[{Williams et~al.(2018)Williams, Nangia, and Bowman}]{Williams2018ABC}
Adina Williams, Nikita Nangia, and Samuel Bowman. 2018.
\newblock \href {https://doi.org/10.18653/v1/N18-1101} {A broad-coverage
  challenge corpus for sentence understanding through inference}.
\newblock In \emph{Proceedings of the 2018 Conference of the North {A}merican
  Chapter of the Association for Computational Linguistics: Human Language
  Technologies, Volume 1 (Long Papers)}, pages 1112--1122, New Orleans,
  Louisiana. Association for Computational Linguistics.

\bibitem[{Zellers et~al.(2019)Zellers, Holtzman, Bisk, Farhadi, and
  Choi}]{Zellers2019HellaSwagCA}
Rowan Zellers, Ari Holtzman, Yonatan Bisk, Ali Farhadi, and Yejin Choi. 2019.
\newblock \href {https://doi.org/10.18653/v1/p19-1472} {Hellaswag: Can a
  machine really finish your sentence?}
\newblock In \emph{Proceedings of the 57th Conference of the Association for
  Computational Linguistics, {ACL} 2019, Florence, Italy, July 28- August 2,
  2019, Volume 1: Long Papers}, pages 4791--4800. Association for Computational
  Linguistics.

\bibitem[{Zhang et~al.(2019)Zhang, Baldridge, and He}]{Zhang2019PAWSPA}
Yuan Zhang, Jason Baldridge, and Luheng He. 2019.
\newblock \href {https://doi.org/10.18653/v1/N19-1131} {{PAWS}: Paraphrase
  adversaries from word scrambling}.
\newblock In \emph{Proceedings of the 2019 Conference of the North {A}merican
  Chapter of the Association for Computational Linguistics: Human Language
  Technologies, Volume 1 (Long and Short Papers)}, pages 1298--1308,
  Minneapolis, Minnesota. Association for Computational Linguistics.

\end{thebibliography}
